%% file: main.tex
\documentclass[10pt,twocolumn,letterpaper]{article}

\usepackage[pagenumbers]{wacv} 

\input{preamble}

\definecolor{wacvblue}{rgb}{0.21,0.49,0.74}
\usepackage[pagebackref,breaklinks,colorlinks,allcolors=wacvblue]{hyperref}

\def\wacvPaperID{3156} 
\def\confName{WACV}
\def\confYear{2026}

\title{MAFM³: Modular Adaptation of Foundation Models for Multi-Modal Medical AI}

\author{
Mohammad Areeb Qazi,
Munachiso S Nwadike,
Ibrahim Almakky,
Mohammad Yaqub,
Numan Saeed \\
Mohamed bin Zayed University of Artificial Intelligence \\
Abu Dhabi, UAE \\
{\tt\small \{mohammad.qazi, munachiso.nwadike, ibrahim.almakky, mohammad.yaqub, numan.saeed\}@mbzuai.ac.ae}
}

\begin{document}
\maketitle
\input{sec/0_abstract}    
\input{sec/1_intro}
\input{sec/2_related_works}
\input{sec/3.0_methodology}
\input{sec/4_experimental_setup}
\input{sec/5_results_and_discussion}

\input{sec/6_conclusion}
{
    \small
    \bibliographystyle{ieeenat_fullname}
    \bibliography{main}
}

\end{document}

%% file: preamble.tex
\usepackage{multirow}
\usepackage{makecell}
\usepackage{pifont}

%% file: sec/0_abstract.tex
\begin{abstract}
Foundational models are trained on extensive datasets to capture the general trends of a domain. However, in medical imaging, the scarcity of data makes pre-training for every domain, modality, or task challenging. Instead of building separate models, we propose MAFM³ (Modular Adaptation of Foundation Models for Multi-Modal Medical AI), a framework that enables a single foundation model to expand into diverse domains, tasks, and modalities through lightweight modular components. These components serve as specialized skill sets that allow the system to flexibly activate the appropriate capability at the inference time, depending on the input type or clinical objective. Unlike conventional adaptation methods that treat each new task or modality in isolation, MAFM³ provides a unified and expandable framework for efficient multitask and multimodality adaptation. Empirically, we validate our approach by adapting a chest CT foundation model initially trained for classification into prognosis and segmentation modules. Our results show improved performance on both tasks. Furthermore, by incorporating PET scans, MAFM³ achieved an improvement in the Dice score 5\% compared to the respective baselines. These findings establish that foundation models, when equipped with modular components, are not inherently constrained to their initial training scope but can evolve into multitask, multimodality systems for medical imaging. The code implementation of this work can be found at \href{https://github.com/Areeb2735/CTscan_prognosis_VLM}{Code}.
\end{abstract}

%% file: sec/1_intro.tex
\section{Introduction}
\label{sec:intro}

Medical imaging plays a crucial role in modern healthcare, providing essential visual insights that are key to diagnosing, monitoring, and treating a wide range of conditions \cite{litjens2017survey,giger2018machine,ranschaert2019artificial}. Despite its importance, collecting and curating medical imaging data is highly resource-intensive, with large hospitals generating approximately 100 terabytes of imaging data per year, which require significant storage, processing power, and infrastructure to effectively manage \cite{duke2015healthdata,langer2011multisite,openmed2023bigdata}. As a result, it becomes difficult to share medical data. Although these challenges limit the feasibility of training distinct models for each task or modality, they also highlight the potential of modular adaptation strategies that can extend a single foundation model to multiple specialized functions without requiring extensive retraining.  

Now, with the introduction of Medical Foundational Models (FM) pre-trained on large-scale data, deep learning systems have demonstrated exceptional performance within their trained domains \cite{zhao2024biomedparse,wu2023radfm,hamamci2024developing,liang20243d,zhang2024foundation}. These models are typically customized to their specific domains by leveraging specialized datasets, architectures, and training objectives that align with medical imaging tasks. For example, they are trained on radiology reports, CT scans, MRIs, or pathology slides, enabling them to capture domain-specific patterns and features that general-purpose models may overlook \cite{wu2023radfm,yu2024chief,taher2023eden}. However, the diversity of medical imaging modalities presents a significant challenge for FM. For example, CT provides detailed cross-sectional images of anatomical structures, while PET visualizes metabolic processes, highlighting functional aspects of tissues. Each modality captures different aspects of human anatomy and physiology, leading to different data distributions. This variability can hinder the generalizability of FM in all types of imaging, as they may struggle to perform consistently well on modalities not represented in their training data \cite{shi2024trustworthy}. However, rather than viewing these models as static systems tied to a narrow scope, they can be seen as flexible backbones that, when equipped with lightweight modular components, can serve as the basis for multitask, multimodal medical AI systems.  

\begin{figure}[t]
    \centering
    \includegraphics[width=\linewidth]{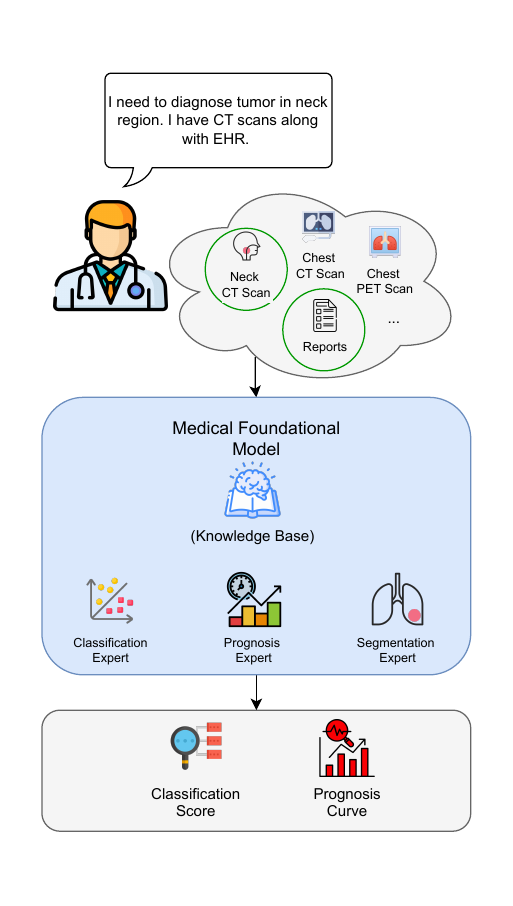}
    \caption{Overview of the objective behind MAFM$^{3}$. A clinician provides diverse medical data sources such as CT scans, PET scans, and electronic health records (EHR). The Medical Foundational Model serves as a central knowledge base, from which task-specific experts (classification, prognosis, segmentation) can be activated depending on the input and clinical requirement. This modular design enables flexible adaptation of a single foundational model to multiple modalities and tasks, supporting generalist medical AI applications.}
    \label{fig:overview_method}
\end{figure}

Training FM for each medical imaging modality is not feasible due to the vast data requirements, which are often unavailable in healthcare. As a result, adaptation techniques have been used to extend models across domains, classes, and tasks \cite{saadi2024pemma,qazi2024dynammo,zhang2023adapter,chen2024lowrankmoe,sun2024continually}. Studies have demonstrated the adaptation of classification models to incorporate new classes while maintaining their original task \cite{qazi2024continual,zhang2023adapter}, segmentation models to generalize across different anatomical regions \cite{chen2024lowrankmoe}, and report generation models to generate structured results for diverse medical findings \cite{sun2024continually}. Furthermore, research has shown that models can be adapted to integrate multiple modalities over the same anatomical region \cite{sobirov2022automatic,saadi2024pemma}.  

Despite these advancements, existing approaches primarily focus on continual adaptation within a single task or domain, lacking a unified framework to extend foundation models across new tasks, imaging modalities, and anatomical regions. Moreover, unlike pre-training on curated datasets, real-world medical imaging data often varies in resolution and quality, posing additional challenges in adapting FM to diverse tasks and domains. Beyond continual learning, a broader perspective is to treat each new module as a capacity unit that can be selectively activated depending on the type of input and the clinical objective. This creates a efficient multitask and multimodal workflow, avoiding the need to train specialized model ensembles and aligning with real-world scenarios where different imaging modalities and tasks coexist.  

To this end, we propose MAFM³ (Modular Adaptation of Foundation Models for Multimodal Medical AI), a framework that enables a single FM initially trained for a specific modality, task, or anatomical region to expand with modular components that specialize in new tasks, modalities, and regions. Unlike conventional approaches to continual learning or domain adaptation, which often focus narrowly on sequential extension, MAFM³ emphasizes modularity and orchestration, allowing the foundation to activate the appropriate component during inference flexibly. This makes the system scalable across diverse adaptation scenarios and efficient and practical for deployment in medical imaging workflows. Figure~\ref{fig:overview_method} illustrates an overview of the proposed framework, showing how a single foundation model can be expanded with modular components and selectively activated to handle various tasks and modalities.

Our contributions can be summarized as follows.  
\begin{itemize}
    \item We introduce MAFM³, a modular framework that enables foundation models to be efficiently extended across new tasks, modalities, and anatomical regions without training separate models.  
    \item We propose a method for adapting foundation models to arbitrary resolution sizes, allowing seamless generalization beyond their original training constraints.  
    \item We show that modular extensions can be selectively activated at inference, enabling an efficient multi-task, multi-modality workflow that reduces redundancy and preserves performance across prior tasks.  
\end{itemize}

%% file: sec/2_related_works.tex
\section{Related Works}

Deep learning has significantly advanced medical imaging by improving diagnostic accuracy and workflow efficiency in various clinical applications~\cite{litjens2017survey,giger2018machine,ranschaert2019artificial}. However, the development of high-performing models often depends on access to large curated datasets, which are challenging to acquire due to storage, privacy, and infrastructure constraints~\cite{duke2015healthdata,langer2011multisite,openmed2023bigdata}. To address these limitations, researchers have explored medical foundation models, continual learning techniques, and more recently modular adaptation strategies. In the following, we summarize the most relevant works.

\subsection{Medical Foundation Models}

Medical foundation models are pre-trained on large-scale datasets to capture generalizable representations transferable across downstream tasks. Examples include BiomedParse~\cite{zhao2024biomedparse}, CHIEF~\cite{yu2024chief}, and EDEN~\cite{taher2023eden}, which leverage high-resolution cross-modal medical data to learn robust feature spaces. These models have demonstrated strong performance in classification, segmentation, and report generation in the radiology and pathology domains~\cite{wu2023radfm,zhang2024foundation,liang20243d}. For example, BiomedParse integrates radiology reports with imaging data for structured representation learning, while CHIEF explores hierarchical embeddings to improve retrieval and diagnostic reasoning. Despite these advances, foundation models are often constrained by the scope of their pre-training data, leading to degraded performance on unseen modalities or anatomical regions~\cite{shi2024trustworthy}. This limitation motivates strategies that allow models to evolve beyond their original training distribution.  

Beyond medical imaging, the broader machine learning community has also developed general-purpose foundation models such as CLIP~\cite{radford2021clip}, Flamingo~\cite{alayrac2022flamingo}, and LLaVA~\cite{liu2023llava}, which align vision and language modalities for multi-task reasoning. Inspired by these works, recent medical models such as BioViL-T~\cite{boecking2022biovil} and MedCLIP~\cite{wang2022medclip} adapt contrastive learning to radiology reports and imaging data, demonstrating that language can provide crucial contextual guidance for medical tasks. However, these approaches typically remain fixed after pre-training and do not directly address the need for continual or modular expansion.

\subsection{Continual Learning in Medical Imaging}

Continual learning (CL) aims to extend the capabilities of a model sequentially without retraining from scratch or forgetting prior knowledge. Previous work in medical imaging has demonstrated the feasibility of incrementally extending models across new classes or domains while preserving earlier performance~\cite{qazi2024continual,zhang2023adapter}. For example, DynaMMo~\cite{qazi2024dynammo} dynamically merges task-specific modules into a shared backbone to enable class-incremental learning with minimal overhead. Similarly, PEMMA~\cite{saadi2024pemma} leverages low-rank adapter tuning to integrate PET imaging into pre-trained CT models, demonstrating that modality fusion can be achieved without catastrophic forgetting. Task adaptation has also been explored: classification models have been extended to segmentation or prognosis tasks with minimal trade-offs in accuracy by introducing lightweight components such as LoRA or adapter layers~\cite{chen2024lowrankmoe,sun2024continually}. These works establish CL as a viable direction for extending foundation models in the medical domain.  

At the same time, limitations persist in scaling CL approaches to heterogeneous multi-modal datasets. Most studies evaluate only a handful of tasks or modalities, and robustness to domain shift or long adaptation sequences remains largely untested. Moreover, performance is rarely benchmarked against strong task-specific baselines under rigorous statistical validation, leaving open questions about reliability in real-world medical settings.

\subsection{Modular Adaptation Strategies}

Beyond continual learning, modular adaptation has emerged as a promising strategy to provide scalable, plug-and-play expansion. Unlike pure CL, which emphasizes sequential training, modular approaches focus on equipping models with specialized components that can be selectively activated at inference. One line of work proposes self-expanding adapters, where model capacity is dynamically increased when distribution shifts are detected, ensuring adaptability without compromising prior tasks~\cite{wang2024sema}. Another approach introduces task-specific adapters that are continually trained and later merged with retrieval mechanisms to preserve knowledge in class-incremental learning~\cite{sun2025mos}. In the medical domain, low-rank mixture-of-experts models have been explored for segmentation, where each expert subnetwork specializes in an organ or lesion class, thereby isolating parameters to mitigate interference~\cite{chen2024lowrankmoe}.  

In the broader machine learning literature, modular designs such as mixture-of-experts transformers~\cite{shazeer2017moe,lepikhin2021gshard} and parameter-efficient fine-tuning methods~\cite{houlsby2019adapter,hu2021lora} have shown that large foundation models can be adapted to new domains without retraining the full network. These ideas are beginning to influence medical imaging research, but systematic exploration of modularity for multi-task, multi-modal, and multi-domain adaptation remains limited.

\subsection{Limitations of Existing Approaches}

Despite these advances, most methods treat adaptation across tasks, modalities, or anatomical regions as isolated processes, limiting the development of truly generalist systems. In practice, medical imaging data are also heterogeneous in resolution, quality, and distribution, further complicating adaptation. Existing frameworks either focus narrowly on continual learning without modular flexibility or on modular expansion without a unified integration across tasks and modalities. Furthermore, very few works explicitly address catastrophic forgetting in the context of modular expansion, or provide quantitative analyses of compute efficiency relative to full fine-tuning.  

Addressing this gap, our framework \textbf{MAFM$^3$} builds on continual learning and modular adaptation by offering a unified and extensible approach. Through modular components that can be selectively activated during inference, MAFM$^3$ enables foundation models to scale across tasks, modalities, and anatomical regions, contributing to domain-robust, multipurpose AI systems for medical imaging.

%% file: sec/3.0_methodology.tex
\section{Methodology}

\begin{figure*}[t]
    \centering
    \includegraphics[width=\linewidth]{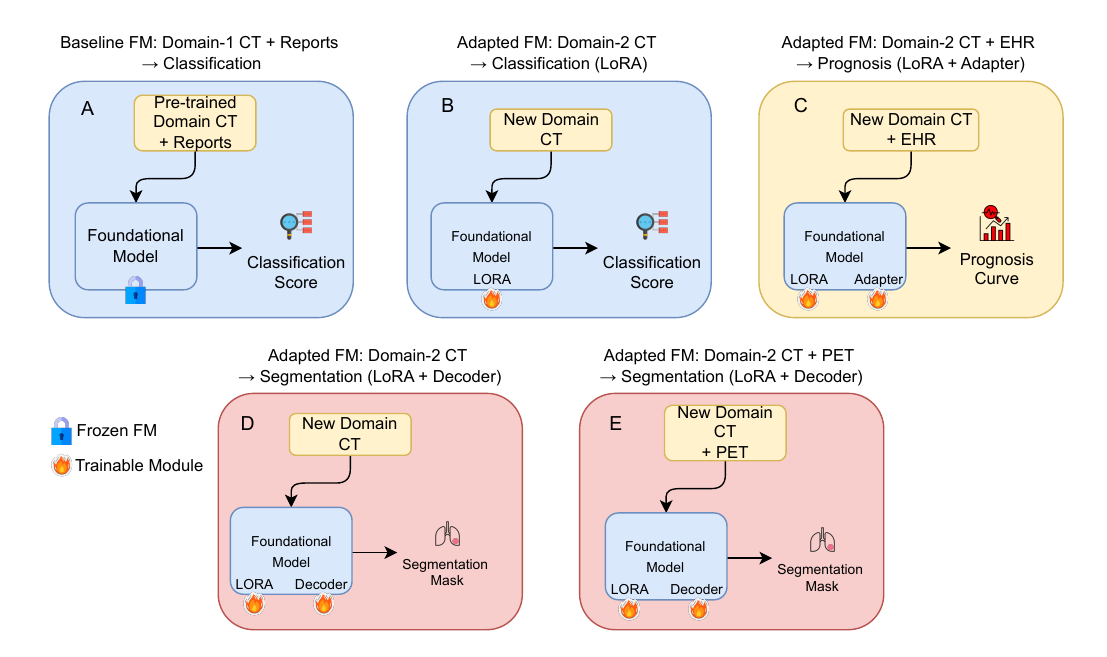}
    \caption{Conceptual overview of MAFM$^{3}$. A single foundational model trained on Domain-1 CT+Reports for classification is adapted with lightweight modules (LoRA, adapters, decoder) for new domains, inputs, and tasks. Each panel illustrates an independent adaptation scenario, highlighting the modular extensibility of the framework.}
    \label{fig:overall_method}
\end{figure*}

\subsection{Problem Statement} 

Clinical requirements are highly dynamic in nature. A doctor may begin with the need for CT-based disease classification, later demand prognosis prediction incorporating reports, and eventually add PET scans for segmentation. It is not feasible to retrain a new model from scratch each time a requirement changes. Instead, what is needed is a framework that can modularly adapt to evolving demands—extending capabilities across modalities (CT, PET, reports), tasks (classification, prognosis, segmentation), and anatomical regions—while preserving the foundational knowledge of the base model.  

Formally, let $F_0$ denote a foundation model with frozen encoders for images and text. The goal is to adapt this model to new datasets $\{X_i\}_{i=1...M}$ and support up to $N$ tasks $\{Y_j\}_{j=1...N}$, each with its own output space. For example, $Y_{1}$ could be prognosis labels derived from reports $X_{5}$, while $Y_{2}$ could be segmentation masks generated from PET scans $X_{42}$. The guiding principle is that a clinician should be able to provide any arbitrary set of inputs and request any arbitrary set of outputs.  

We define MAFM$^3$ abstractly as:
\begin{equation}
F_0 \;\mapsto\; F_A, \quad 
F_A : \{X_i\} \mapsto \{Y_j\}.
\end{equation}

The key point is not simply to reach a final adapted model, but to acknowledge that medical requirements emerge incrementally. The framework must adapt modularly, step by step, without retraining from scratch. This motivates a dynamic view in which the model evolves through a sequence:
\begin{equation}
F_0, \; F_1, \; \ldots, \; F_T,
\end{equation}
where each $F_t$ extends the functionality of the previous model while retaining prior capabilities. After $T$ steps, we obtain:
\begin{equation}
F_T : 
\Bigg(\bigcup_{t=0}^{T} X_i \Bigg) 
\;\mapsto\; 
\Bigg(\bigcup_{t=0}^{T} Y_j \Bigg).
\end{equation}

\noindent\textbf{Illustrative cases.}  
- Base case:  
\begin{equation}
F_0 : \{X_{\text{CT}}, X_{\text{R}}\} \;\mapsto\; \{Y_{\text{Class}}\}.
\end{equation}

- Adding a new dataset (reports) that supports multiple outputs (disease labels and prognosis):  
\begin{equation}
\begin{aligned}
F_1 : & \; \{X_{\text{CT}}, X_{\text{R}}\} 
          \mapsto \{Y_{\text{Class}}, Y_{\text{Prog}}\}, \\
F_1 = & \; F_0 + \{Y_{\text{Prog}}\}.
\end{aligned}
\end{equation}

- Adding multiple new modalities (PET, MRI) that support a single shared output (segmentation):  
\begin{equation}
\begin{aligned}
F_2 : & \; \{X_{\text{CT}}, X_{\text{R}}, X_{\text{PET}}, X_{\text{MRI}}\} \\
      \mapsto & \; \{Y_{\text{Class}}, Y_{\text{Prog}}, Y_{\text{Seg}}\}, \\
F_2 = & \; F_1 + \{Y_{\text{Seg}}\}.
\end{aligned}
\end{equation}

These examples illustrate that adaptation steps need not be one-to-one: a single dataset may unlock multiple outputs, while multiple modalities may serve a single task. The core property is cumulative growth: once a foundation model acquires a skill, later extensions preserve it while adding new ones.  

\subsection{Selected Clinical Applications}

To illustrate the framework, we demonstrate MAFM$^3$ on three representative medical tasks: classification, prognosis prediction, and segmentation. These span distinct output types (categorical labels, temporal outcomes, voxel-wise masks) and heterogeneous inputs (CT, PET, reports), providing a broad yet tractable test of adaptability. The emphasis is on demonstrating incremental growth: if $F_n$ works, then so should $F_{n+1}$.

\subsubsection{Classification ($F_0$)}
Disease classification serves as the baseline capability. Inputs are CT scans and radiology reports, and outputs are categorical disease labels:
\begin{equation}
F_0 : \{X_{\text{CT}}, X_{\text{R}}\} \;\mapsto\; \{Y_{\text{Class}}\}.
\end{equation}

\subsubsection{Prognosis ($F_1$)}
Prognosis prediction estimates disease progression and outcomes, requiring images and a textual context. The challenge is to add this capability without degrading classification. Formally:
\begin{equation}
\begin{aligned}
F_1 : & \; \{X_{\text{CT}}, X_{\text{PET}}, X_{\text{R}}\} \\
      \mapsto & \; \{Y_{\text{Class}}, Y_{\text{Prog}}\}, \\
F_1 = & \; F_0 + \{Y_{\text{Prog}}\}.
\end{aligned}
\end{equation}

\subsubsection{Segmentation ($F_2$)}
Segmentation requires voxel-level reconstruction and multimodal fusion. Extending $F_1$:
\begin{equation}
\begin{aligned}
F_2 : & \; \{X_{\text{CT}}, X_{\text{PET}}, X_{\text{R}}\} \\
      \mapsto & \; \{Y_{\text{Class}}, Y_{\text{Prog}}, Y_{\text{Seg}}\}, \\
F_2 = & \; F_1 + \{Y_{\text{Seg}}\}.
\end{aligned}
\end{equation}

\subsubsection{General Extensibility}
The principle generalizes: each new $F_t$ extends $F_{t-1}$ with new inputs or outputs while retaining prior ones:
\begin{equation}
F_t = F_{t-1} + \{\text{new inputs, new outputs}\}.
\end{equation}
By induction:
\begin{equation}
F_n : 
\Bigg(\bigcup_{t=0}^n X_i\Bigg) 
\;\mapsto\; 
\Bigg(\bigcup_{t=0}^n Y_j\Bigg).
\end{equation}

Thus, MAFM$^3$ is not limited to three tasks but can flexibly integrate arbitrary new modalities (e.g., MRI, genomics) and tasks (e.g., new prognostic markers) without retraining from scratch.

\subsection{Modular Adaptation Mechanisms in MAFM$^3$}

Modular adaptation in MAFM$^3$ is realized through these complementary mechanisms:  

\paragraph{I. Within-model Adaptation (WMA).}  
Parameter-efficient fine-tuning is performed using LoRA~\cite{hu2021lora}. Given a frozen weight matrix $W$, we learn:
\begin{equation}
W' = W + \Phi_1 \Phi_2,
\end{equation}
where $\Phi_1 \in \mathbb{R}^{d \times r}$, $\Phi_2 \in \mathbb{R}^{r \times h}$, and $r \ll \min(d,h)$.

\paragraph{II. Post-model Adaptation (PMA).}  
Lightweight MLP layers refine extracted features for task-specific outputs. For segmentation, a decoder reconstructs voxel-wise masks. For multimodal inputs, fusion MLPs integrate CT, PET, and report embeddings into a shared latent space.

\paragraph{III. Resolution Adaptation.}  
Since medical scans vary in resolution across institutions, task-specific patch embedding layers $P: \mathbb{R}^{H \times W \times C} \to \mathbb{R}^d$ and fine-tuned positional embeddings allow the backbone to adapt seamlessly to heterogeneous inputs.

Together, these mechanisms ensure that each extension $F_t$ equals $F_{t-1}$ plus new capabilities. The core property is cumulative growth without forgetting. Once a medical foundation model acquires a skill such as classification, it retains it even as prognosis and segmentation are added. This modular extensibility ensures that the system evolves with clinical needs rather than being retrained from scratch for every new requirement.

%% file: sec/4_experimental_setup.tex
\section{Experimental Setup}

\input{tables/experiment_tables}

\subsection{Dataset Description} 
We use the publicly available HECTOR dataset (HEad and neCK TumOR)~\cite{andrearczyk2021overview}, one of the few resources that provides multiple modalities in a single benchmark. The dataset comprises CT and PET scans, segmentation masks, and electronic health records (EHR) of 488 patients collected at seven centers with heterogeneous scanner types. Importantly, HECTOR includes Recurrence-Free Survival (RFS) information, providing both time-to-event outcomes and censoring status, making it well suited for evaluating both prognosis and segmentation tasks in a unified setting.

\subsection{Foundation Model Setup} 
We employ CT-CLIP~\cite{hamamci2024developing} as our foundational model. CT-CLIP is pre-trained in chest CT scans for classification, utilizing contrastive learning between images and texts. This pretrained model serves as the frozen backbone in our experiments, which we then extend modularly across tasks and modalities following the progressive sequence in Table~\ref{tbl:exp_main}. This progressive adaptation is illustrated in Figure~\ref{fig:overall_method}, which shows how the foundational baseline model trained in Domain-1 CT + Reports (panel A) is incrementally extended for classification, prognosis, and segmentation across new domains and modalities (panels B-E).
Specifically, the foundation is first adapted for different domain CT classification, then for prognosis prediction, then for CT scan segmentation, and finally for multimodal segmentation using both CT and PET.

\subsection{Task-Specific Configurations} 

\subsubsection{Prognosis Prediction}
For the prediction of patient outcomes, we integrate two survival models: DeepHit~\cite{lee2018deephit} and MTLR~\cite{fotso2018deep}. Latent embeddings from CT-CLIP are enriched with within-model (LoRA) and post-model (MLP) adapters before being passed to the survival models. This approach minimizes parameter overhead while enabling task-specific fine-tuning. As shown in Figure~\ref{fig:overall_method} (Panel C), prognosis prediction is instantiated by enriching the foundational model with LoRA and adapter modules to integrate CT and EHR data. 

\noindent\textbf{Preprocessing.} CT scans are resized at CT-CLIP resolution, cut to range $(-1024, 1024)$, and normalized to $[-1, 1]$. The EHR features (e.g. age, weight, gender) are reformulated into structured textual reports using GPT-4 instructions~\cite{achiam2023gpt}, following a consistent template to ensure standardization.  

\noindent\textbf{Training.} Models are trained for 50 epochs with AdamW (learning rate $3 \times 10^{-4}$, weight decay $1 \times 10^{-5}$, batch size 16). The best checkpoint is selected based on the highest validation concordance index (C-index).  

\subsubsection{Segmentation}
For segmentation, we use the UNETR 3D model~\cite{Hatamizadeh_2022} as a baseline and adapt it within the MAFM$^3$ framework. As shown in Figure~\ref{fig:overall_method}, Panels D–E, the foundational model is extended with LoRA weights and a task-specific decoder. For multimodal segmentation, PET scans are introduced via modality-specific tokens fused with CT inputs.

\noindent\textbf{Preprocessing.} CT and PET volumes are resized to $96^3$, clipped and normalized; PET images are additionally Z-score normalized.  

\noindent\textbf{Training.} All models are trained for a maximum of 25k steps using the MONAI library~\cite{cardoso2022monai}. We employ AdamW (learning rate $1 \times 10^{-4}$, weight decay $1 \times 10^{-5}$, batch size 1). For CT-only segmentation, LoRA weights and a decoder are added to the foundation model. For CT+PET segmentation, modality-specific tokens are introduced and fused via a dedicated adapter. The best checkpoint is selected based on the highest validation Dice score.  

%% file: tables/experiment_tables.tex

\begin{table*}[!t]
\centering
\caption{Progressive adaptation learning sequence in our experiments.}
\label{tbl:exp_main}
\setlength{\tabcolsep}{4pt}
\scalebox{0.8}{
\begin{tabular}{l | c c c c}
\toprule
{\textbf{Order}$\rightarrow$} & {\textbf{Foundational Model}} & {\textbf{Step 1}} & {\textbf{Step 2}} & {\textbf{Step 3}} \\
\midrule
\multirow{1}{*}{\textbf{Body Region}} 
 & Chest $\rightarrow$ & Head \& Neck $\rightarrow$ & Head \& Neck & $\rightarrow$ Head \& Neck \\
\multirow{1}{*}{\textbf{Modality}} 
 & CT $\rightarrow$ & CT $\rightarrow$ & CT $\rightarrow$ & CT \& PET \\
\multirow{1}{*}{\textbf{Task}} 
 & Classification $\rightarrow$ & Prognosis $\rightarrow$ & Segmentation $\rightarrow$ & Segmentation \\
\bottomrule
\end{tabular}}
\end{table*}

%% file: sec/5_results_and_discussion.tex
\section{Results and Discussion}

\input{tables/main_tables}

\subsection{Prognosis Prediction Results}

We first evaluate the prognostic performance of our adapted CT-CLIP model on the HECKTOR dataset using 5-fold cross-validation. Prognosis prediction is a particularly challenging task, as it requires integrating multimodal information and modeling long-term patient outcomes. Baseline performance is established using two widely used survival models—DeepHit~\cite{lee2018deephit} and MTLR~\cite{fotso2018deep}.  

Initially, we adapt the CT-CLIP model for the prognosis task by incorporating both within-model adaptation (LoRA) and post-model adaptation (MLP layers) applied to the text encoder. As shown in Table~\ref{tbl:prognosis_main}, these modular extensions consistently improve performance compared to the baseline task-specific models. Specifically, we observe an approximate 2\% improvement in the C-index over standard baselines, demonstrating that foundation model embeddings can be effectively repurposed for outcome prediction when enriched with lightweight modules.  

Importantly, the best results are achieved when adaptation methods are applied simultaneously. Adapting both the text and image encoders using LoRA and MLP adapters with DeepHit yields the highest C-index of 0.721, outperforming the baseline by a clear margin. This underscores the benefit of leveraging multimodal information—particularly textual EHR data, which provides contextual details not captured in imaging alone. Interestingly, adapting only the image encoder did not produce significant improvements, suggesting that prognosis prediction in this dataset relies more heavily on structured clinical features embedded in the reports. This observation highlights the complementary value of textual modalities in prognostic modeling and validates the multimodal design of MAFM$^3$.

\subsection{Segmentation Results}

We next examine segmentation, another clinically important task, as precise delineation of anatomical structures and tumors underpins treatment planning, radiotherapy, and quantitative disease assessment. For this experiment, we trained UNETR~\cite{Hatamizadeh_2022} as the baseline and compared it to our MAFM$^3$-adapted versions.  

The adaptation was performed in two stages. First, we adapted the foundational model for CT-only segmentation by introducing a task-specific decoder (post-model adaptation). Table~\ref{tbl:segmentation_main} shows that this adaptation yielded a Dice score improvement of approximately 2\% compared to the baseline. Second, we extended the framework to multimodal segmentation by incorporating PET scans alongside CT. This multimodal configuration produced the highest Dice score of 65.7\%, clearly outperforming both the baseline and the CT-only adaptation. These results demonstrate that MAFM$^3$ can flexibly extend foundation models to voxel-level prediction tasks, while efficiently incorporating complementary modalities such as PET to boost segmentation accuracy.

\subsection{Framework-Level Insights}

Taken together, these results illustrate how MAFM$^3$ enables foundation models to progressively extend across tasks and modalities in a modular, scalable fashion. Instead of retraining a separate model for each new task, our approach reuses the frozen CT-CLIP backbone as a knowledge base and activates lightweight modules for specialization. This validates the central principle of the framework: that once a model acquires a capability (e.g., classification), it can retain it while new capabilities (e.g., prognosis, segmentation) are added incrementally.  

From a clinical perspective, even modest improvements in metrics can be impactful. For prognosis prediction, an increase in 2\% the C index can translate into a more reliable patient stratification for treatment planning. For segmentation, improved Dice scores directly affect the accuracy of tumor delineation, which has downstream implications in radiotherapy dose calculations. Thus, beyond statistical gains, the framework contributes meaningfully to real-world clinical workflows.

\subsection{Efficiency and Robustness}

Another important observation is the efficiency of the modular adaptation process. By fine-tuning only a small subset of parameters (via LoRA and lightweight adapters), the foundational model is extended without retraining or overwriting its original knowledge. This significantly reduces computational costs compared to the full model fine-tuning or ensemble methods. Although we did not explicitly measure FLOPs or memory savings in this work, the reduction in trainable parameters alone suggests substantial efficiency gains, making the approach well suited for healthcare settings where compute resources are limited.  

In addition to efficiency, the framework demonstrates robustness to domain shift. Although CT-CLIP was originally pre-trained on chest CT scans, our results show that it can be extended to head-and-neck imaging and multimodal CT+PET tasks with competitive performance. This indicates that the learned representations transfer well across anatomical regions and modalities when coupled with modular adaptation. By decoupling the frozen backbone from lightweight task-specific modules, MAFM$^3$ avoids catastrophic forgetting and supports incremental growth, even in heterogeneous settings.

In summary, the experiments confirm that MAFM$^3$ provides a flexible and generalizable approach to extend the foundation models in medical imaging. The prognostic and segmentation tasks, which represent distinct outputs and input configurations, were successfully triggered without retraining the backbone. Performance improvements over baselines, efficient parameter usage, and robustness to new domains collectively highlight the promise of modular adaptation. At a higher level, the results support our central claim. Medical foundation models need not be static but can evolve dynamically to meet the ever-changing demands of clinical practice.


Although our study demonstrates the promise of MAFM$^3$ for modular adaptation of medical foundation models, several limitations remain. First, our experiments are restricted to the HECKTOR dataset, which, although multimodal, represents a single clinical domain. The generalizability of the framework to other anatomical regions such as chest or brain, and to additional modalities such as MRI, histopathology, or genomics, has not yet been empirically validated. Second, we evaluated only three representative tasks, classification, prognosis prediction, and segmentation, which, while diverse, do not capture the full range of clinical needs such as detection, registration, treatment planning, or report generation.   

Finally, the robustness of the framework under conditions of extreme data scarcity, noisy EHR inputs, or domain shifts across scanners and institutions was not fully analyzed. Clinical integration will also require additional considerations such as interpretability, uncertainty quantification, and workflow integration, which remain outside the scope of this work.

\section{Future Directions}

Building on the present findings, several avenues of research can extend the utility of MAFM$^3$. One immediate direction is to validate the framework on larger and more diverse datasets covering multiple anatomical regions and imaging modalities such as MRI or pathology slides. This would test the generalizability of the approach beyond head-and-neck oncology. In addition, expanding the set of tasks to include detection, report generation, and treatment response prediction would provide a more comprehensive demonstration of modular adaptability in clinical workflows.  

A second direction is to explicitly analyze catastrophic forgetting under longer adaptation sequences. Future work could design controlled experiments where the framework is adapted across many tasks and domains in sequence, while monitoring the retention of earlier capabilities. This would provide stronger guarantees of robustness and enable comparisons to established continual learning benchmarks. Statistical validation of performance gains, including confidence intervals and hypothesis testing, should also be incorporated to ensure rigor in reporting.  

Another promising extension is the development of automated routing strategies for inference. Rather than manually selecting modules based on the known task and modality, meta-learning or controller networks could dynamically activate the relevant components, making the system more flexible for deployment in real-world settings where inputs are heterogeneous and partially missing. Alongside this, a quantitative analysis of computational efficiency in terms of FLOPs, memory, and training time would substantiate claims of efficiency compared to baselines such as full fine-tuning or ensembles.  

Finally, moving closer to clinical translation will require integrating interpretability methods, uncertainty estimation, and calibration strategies to make outputs more trustworthy for clinicians. Incorporating semi-supervised or self-supervised adaptation could also improve robustness under limited or noisy data. Taken together, these directions can establish MAFM$^3$ not only as a proof-of-concept framework but as a practical tool for scalable, modular, and trustworthy medical AI.

%% file: tables/main_tables.tex
\begin{table}[!t]
\centering
\caption{Prognosis adaptation performance with different input modalities using C-Index. We used DeepHit \cite{lee2018deephit} and MTLR \cite{fotso2018deep} survival model performance under different input modalities.}
\label{tbl:prognosis_main}
\setlength{\tabcolsep}{4pt}
\scalebox{0.85}{
\begin{tabular}{l | c c | c c}
\toprule
{} & \multicolumn{2}{c}{\textbf{Adaptation}} & \multicolumn{2}{c}{\textbf{C-Index}} \\
\multirow{-2}{*}{\makecell[l]{\textbf{Input Modality} $\downarrow$}} & \textbf{PMA} & \textbf{WMA} & \textbf{MTLR} & \textbf{DeepHit} \\
\midrule
\multirow{1}{*}{\textbf{Baseline \cite{saeed2024survrnc}}} 
 & \ding{55} & \ding{55} & 0.634 & 0.661 \\
\midrule
\multirow{2}{*}{\textbf{Text}}
 & \ding{51} & \ding{55} & 0.668 & 0.686 \\
 & \ding{51} & \ding{51} & 0.652 & 0.683 \\
\midrule
\multirow{2}{*}{\textbf{Image}} 
 & \ding{51} & \ding{55} & 0.546 & 0.566 \\
 & \ding{51} & \ding{51} & 0.603 & 0.626 \\
\midrule
\multirow{2}{*}{\textbf{Image and Text}}
 & \ding{51} & \ding{55} & 0.658 & 0.698 \\
 & \ding{51} & \ding{51} & 0.670 & 0.721 \\
\bottomrule
\end{tabular}}
\end{table}

\vspace{1em}

\begin{table}[!t]
\centering
\caption{Performance of Segmentation Adaptation with Different Input Medical Modalities in Dice Score (D-Score). We used UnetR \cite{Hatamizadeh_2022} model for baseline and adapted it for the experiments.}
\label{tbl:segmentation_main}
\setlength{\tabcolsep}{4pt}
\scalebox{0.85}{
\begin{tabular}{l | c c | c}
\toprule
{} & \multicolumn{2}{c}{\textbf{Adaptation}} & {} \\
\multirow{-2}{*}{\makecell[l]{\textbf{Medical Modality} $\downarrow$}} & \textbf{PMA} & \textbf{WMA} & \multirow{-2}{*}{\makecell[l]{\textbf{D-Score}}} \\
\midrule
\multirow{1}{*}{\textbf{CT (Baseline) \cite{Hatamizadeh_2022}}} 
 & \ding{55} & \ding{55} & 0.609\\
\midrule
\multirow{2}{*}{\textbf{CT}} 
 & \ding{51} & \ding{55} & 0.615 \\
 & \ding{51} & \ding{51} & 0.628 \\
\midrule
 \multirow{1}{*}{\textbf{CT \& PET}}
 & \ding{51} & \ding{51} & 0.657 \\
\bottomrule
\end{tabular}}
\end{table}

%% file: sec/6_conclusion.tex
\section{Conclusion}


In this work, we introduced MAFM$^3$, a modular adaptation framework designed to extend the capabilities of medical foundation models across multiple domains, modalities, and tasks. Unlike conventional approaches focusing on isolated adaptation, MAFM$^3$ provides a unified mechanism for expanding foundational knowledge while preserving previously acquired skills. Through experiments on the HECKTOR dataset, we demonstrated that the framework enables effective adaptation for classification, prognosis prediction, and segmentation with minimal computational overhead, while achieving performance competitive with task-specific baselines.  

This work highlights the potential of modular adaptation strategies to address key challenges in medical AI, including data scarcity, domain variability, and the need for scalable generalist systems. By fine-tuning only lightweight modules rather than retraining entire models, MAFM$^3$ lowers the barrier to applying foundation models to diverse clinical settings. A remaining challenge is to quantify the data requirements for adaptation systematically and assess robustness under more extensive task sequences and heterogeneous modalities. Future research will focus on validating the framework in broader datasets and clinical applications, integrating automated module selection in inference, and incorporating interpretability and uncertainty estimation for real-world use. Ultimately, we envision MAFM$^3$ as a step toward practical, trustworthy, and extensible medical AI systems that can continuously evolve alongside the dynamic demands of clinical practice.